\documentclass[runningheads]{llncs}

\usepackage{graphicx}
\usepackage{adjustbox}
\usepackage{epstopdf}

\usepackage[ruled,vlined,linesnumbered,lined]{algorithm2e}
\usepackage{float}
\usepackage{pgf, subcaption}
\usepackage[numbers]{natbib}
\usepackage{tikz}
\usetikzlibrary{shapes,arrows.meta}
\tikzset{%
  >={Latex[width=2mm,length=2mm]},
            base/.style = {rectangle, rounded corners, draw=black,
                           minimum width=4cm, minimum height=1cm,
                           text centered, font=\sffamily},
  activityStarts/.style = {trapezium, fill=blue!30},
       startstop/.style = {base, fill=red!30},
       decision/.style = { diamond, draw=blue, fill=blue!20,
                        text width=5em, text centered,
                        inner sep=1pt, rounded corners },
    activityRuns/.style = {trapezium, trapezium left angle=70, trapezium right angle=110,draw=black,minimum height=0.5cm, fill=green!30},
         process/.style = {base, minimum width=2.5cm, fill=orange!15,
                           font=\ttfamily},
}

\begin{document}

\title{Triclustering of Gene Expression Microarray Data Using Coarse-Grained Parallel Genetic Algorithm}
%
%\titlerunning{Abbreviated paper title}
% If the paper title is too long for the running head, you can set
% an abbreviated paper title here
%
\author{Shubhankar Mohapatra\inst{}\and
Moumita Sarkar\inst{}\and
Anjali Mohapatra\inst{}\and
Bhawani Sankar Biswal\inst{}}
\authorrunning{Shubhankar et al.}
\titlerunning{Triclustering of gene expression microarray data using CGPGA}
% First names are abbreviated in the running head.
% If there are more than two authors, 'et al.' is used.
%
\institute{DST-FIST Bioinformatics Laboratory, IIIT Bhubaneswar, India
\email{B114042@iiit-bh.ac.in},\email{B114066@iiit-bh.ac.in},\\ \email{anjali@iiit-bh.ac.in},\email{c114002@iiit-bh.ac.in}}
\maketitle              % typeset the header of the contribution
\begin{abstract}
Microarray data analysis is one of the major area of research in the field computational biology. Numerous techniques like clustering, biclustering are often applied to microarray data to extract meaningful outcomes which play key roles in practical healthcare affairs like disease identification, drug discovery etc. But these techniques become obsolete when time as an another factor is considered for evaluation in such data. This problem motivates to use triclustering method on gene expression 3D microarray data. In this article, a new methodology based on coarse-grained parallel genetic approach is proposed to locate meaningful triclusters in gene expression data. The outcomes are quite impressive as they are more effective as compared to traditional state of the art genetic approaches previously applied for triclustering of 3D GCT microarray data.

\end{abstract}

\section{Introduction}\label{sec:intro}

\par
In microarray research, finding groups of genes exhibiting similar expressions, clustering and biclustering  techniques are more commonly used in gene expression analysis~\cite{rubio2008classification},~\cite{hartigan1972direct}. However,these techniques become inefficient when the influence of the time as a factor affects the behavior of expression profiles  ~\cite{gomez2011pattern}. Now, these types of longitudinal experiments are gaining interest in various areas of molecular activities where the evaluation of time is essential. For example, in cell cycles, the evolution of diseases or development at the molecular level is time based as they consider time an important factor of evaluation~\cite{bar2004analyzing}. Hence, triclustering appears to be a valuable mechanism as it allows evaluation of the expression profiles under a block of conditions along with under a subset of time points.

\begin{figure}
\centering
\includegraphics[height=1.5in, width=0.3\textwidth]{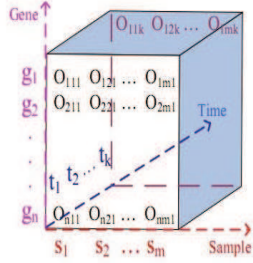}
\caption{Illustration of a tricluster}
\label{fig:tric}
\end{figure}
 
\par
A coherent tricluster is defined as a set of genes that pursues either coherent values or behaviors. These clusters might have useful information that identify significant phenotypes or potential genes relating to the phenotypes and their regulation relations~\cite{tchagang2012mining}.
The computational complexity of triclustering algorithms is more expensive than the biclustering algorithms(which are already NP hard), so heuristic based algorithms are an upstanding resemblance for triclustering.
\par
Genetic Algorithms (GAs) are search specific algorithms and are motivated by the characteristics of genetics and natural selection~\cite{holland1989genetic}. GAs usually undergo some important phases like reproduction, mutation, fitness evaluation and selection. Sequential GAs are competent in many applications as well as in different domains. However, there exist some problems in their utilization of problems like triclustering. For example, the fitness evaluation in sequential GAs is usually very time-consuming. Also, sequential GAs may get trapped in a sub-optimal region of the search space thus becoming unable to find better quality solutions. So parallel GAs(PGAs) seem to be a better alternative to the traditional sequential GAs with the adoption of parallelism. The static subpopulations with migration parallel GAs have a key characteristic of applying multiple demes along with the presence of a migration operator. Coarse-grained parallel genetic algorithms(CgPGA) follow the same general terms for a subpopulation model having a fairly small number of demes with many individuals. Very often coarse-grained parallel GAs are treated as distributed GAs as in general their implementation is carried out on distributed memory MIMD computers. This appeal can also be well configured with heterogeneous networks.
\par
In this paper, an algorithm based on coarse grained parallel genetic algorithms(CgPGA) approach is proposed. This algorithm finds genus of similar patterns for genes on a three-dimensional space, where genes, conditions and time factor are taken into consideration.

\par
%  The key contributions of the proposed frameworks are as follows:
% \begin{itemize}
% \item To use the parallel genetic algorithms(PGAs) in the analysis of gene expression profiles in triclustering domain.
% \item Next, to utilize a static sub-population with migration based parallel genetic approach, namely, coarse-grained algorithm (CgPGA) to select the best individuals.
% \end{itemize} 
% The outcomes of the proposed triclustering algorithm is evaluated in terms of fitness value and computational time with different generations. Further the results are validated with  Gene Ontology(GO) terms to provide functional annotations for the genes.

\par 
The rest of this paper is organized as follows: A review of the literature is presented in section~\ref{rw}. The proposed methodologies along with the details of the fitness functions and the genetic operators used are described in section~\ref{pm}. The simulation results with their GO term validation are discussed in section~\ref{go}. Finally, section~\ref{conc} presents the summary and the research findings of the proposed scheme and prospects for future work.

\section{Related Work}\label{rw}
\par
Zhao and Zaki introduced triCluster algorithm in 2005~\cite{zhao2005tricluster}. In this work, the patterns are discovered in three dimensional (3D) gene expression data along with a set of matrices for the quality measure. A contemporary approach that finds coherent triclusters which contain the regulatory relationships among the genes is stated in~\cite{yin2007mining} and subsequently extract time-delayed clusters in~\cite{wang2010efficiently}.
\par
LagMiner, in~\cite{xu2009finding} introduced a new technique to detect time-lagged 3D clusters. The evolutionary computation in the form of a multi-objective algorithm has also been employed in the search for triclusters in~\cite{liu2008multi}. Bhar Anirban et al. in 2012 presented $\delta$-TRIMAX algorithm~\cite{bhar2012delta}. Again in 2013, the same authors applied the $\delta$-TRIMAX algorithm in estrogen-induced breast cancer cell datasets which provides insights into breast cancer prognosis~\cite{bhar2013coexpression}. David et al. presented a novel tricluster algorithm called as trigen in 2013~\cite{gutierrez2014trigen}. The novelty of this Trigen algorithm lies upon the use of the genetic approach to mine three dimensional gene expression microarray data. In 2015, Ayangleima et al. applied coarse-grained parallel genetic algorithm(CgPGA) with migration technique to mine biclusters in gene expression microarray data~\cite{laishram2015bi}. In the year 2016 Kakati et al. presented a fast gene expression analysis that uses distributed triclustering and parallel biclustering approach~\cite{kakati2016fast}. In her work, the initial bicluster finding is performed by parallel or shared memory approach and then the triclusters are extracted by a distributed or a shared nothing approach. Premalatha et al. in 2016 presented TrioCuckoo~\cite{swathypriyadharsinitriocuckoo} which implemented triclustering using the famous cuckoo search technique.

\section{Proposed Methodology}\label{pm}
\par 
In this section, the reported algorithm has experimented on the standard  yeast cell cycle dataset (Saccharomyces cerevisiae)~\cite{spellman1998comprehensive}. Then the biological validation process is initiated with a tool called GO term finder (Version 0.83)~\cite{boyle2004go} to get the functional annotations of the genes resulted in the output tricluster.

\subsection{\textbf{Encoding of individuals}}\label{eoi}
\par
Every individual in the population encodes a tricluster. Triclusters are represented in the form binary strings of G+C+T length, G being the genes(rows), C being the conditions(columns) and T being the times(height) of the 3D expression matrix. If the bit in an individual is 1, it indicates that the respective row, column or height have a place in the tricluster. 
\subsection{\textbf{Fitness Function}}\label{ff}
\par
Here a fitness function has been implemented  to select the best candidates, which is conceptualized up on the three dimensions aspect of the mean square residue measure (MSR) which has been an all-time effective biclustering measure for gene expression analysis~\cite{cheng2000biclustering}. It is named as $F_{msr}$ now onwards. As $F_{msr}$ is a minimization function, we expect better results with smaller values.

$$F_{msr}(T_C)= MSR- Weights-Distinction$$

The function is defined for every tricluster(TC). It is minimizing and thus lower values are favourable. 
Where,
\subsection{Weights}\label{wts}
\par
The weights term is defined as: 
$$Weights=G_l * w_g + C_l * w_c + T_l *w_t $$
Where $w_g$,$w_c$ and $w_t$ are weights for the number of genes, conditions and times in a  tricluster solution, respectively. High values of weights are favorable.
\subsection{Distinction}\label{dstn}
\par
The distinction term is defined as: 
$$Distinction=CDN_g / G_l *wd_g  + CDN_c / C_l *wd_c + CDN_t / T_l *wd_t.$$
Where,
 $CDN_g$ (Co-ord Distinction no. of g), $CDN_c$ (Co-ord Distinction no. of c) and $CDN_t$ (Co-ord Distinction no. of t) are, respectively, the number of genes, conditions and time coordinates in the tricluster that are absent in the tricluster being evaluated, and $wd_t$,$wd_g$ and $wd_c$ are the distinction weights of the genes, conditions and times respectively. Distinction is a measure for the uniqueness of the tricluster being currently evaluated. With increased value of distinction non-overlapping solutions compared with results previously found can be found. Where,
\begin{itemize}
	\item $G$: Tricluster gene coordinates subset.
	\item $C$ Tricluster condition coordinates subset.
	\item $T$: Tricluster time coordinates subset.
	\item $T_l$: No. of time co-ord of the tricluster
	\item $C_l$: No. of condition co-ord of the tricluster.
	\item $G_l$: No. of gene co-ord of the tricluster.
	\item $TC_
	v(t,g,c)$: Expression value of gene g under condition c at time t from the expression matrix.
\end{itemize}

\subsection{\textbf{Tri-CgPGA}}\label{tricgpga}

Tri-CgPGA is based on coarse grained genetic algorithms which come under Parallel Genetic Algorithm family. So like coarse grained algorithms, this evolutionary algorithm takes several steps to execute which are illustrated in the flowchart and pseudo-code below.
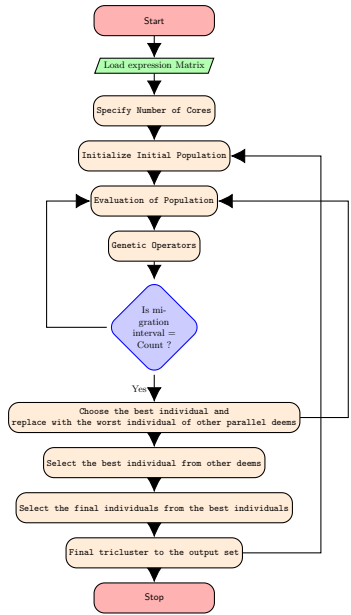
\begin{figure}
\begin{center}

\begin{tikzpicture}[node distance=1.5cm,
    every node/.style={fill=white, font=\sffamily}, align=center,scale=0.4, every node/.style={scale=0.4}]
  % Specification of nodes (position, etc.)
  \node (start)             [startstop]              {Start};
  \node (in1)      [activityRuns, below of=start]
                                                      {Load expression Matrix};
  \node (pr1)     [process, below of=in1]          {Specify Number of Cores};
  \node (pr2)     [process, below of=pr1]          {Initialize Initial Population};
  \node (pr3)     [process, below of=pr2]          {Evaluation of Population};
  \node (pr4)     [process, below of=pr3]          {Genetic Operators};
  \node (dec1)     [decision, below of=pr4,yshift=-1.2cm]          {Is migration interval = Count ?};
  \node (pr5)     [process, below of=dec1,yshift=-1.5cm]          {Choose the best individual and \\replace with the worst individual of other parallel deems};
  \node (pr6)     [process, below of=pr5]          {Select the best individual from other deems};
  \node (pr7)     [process, below of=pr6]          {Select the final individuals from the best individuals};
  \node (pr8)     [process, below of=pr7]          {Final tricluster to the output set};
  \node (stop)     [startstop, below of=pr8]          {Stop};

  \draw[->]             (start) -- (in1);
  \draw[->]             (in1) -- (pr1);
  \draw[->]             (pr1) -- (pr2);
  \draw[->]             (pr2) -- (pr3);
  \draw[->]             (pr3) -- (pr4);
  \draw[->]             (pr4) -- (dec1);
  \draw[->]             (dec1) -- node[xshift=-0.5cm]{Yes} (pr5);
  \draw[->] (dec1.west) -- ++(-2,0) -- ++(0,3.2) -- ++(0,1) --                (pr3.west);
  \draw[->] (pr5.east) -- ++(1.6,0) -- ++(0,6.2) -- ++(0,1) --                (pr3.east);
  \draw[->]             (pr5) -- (pr6);
  \draw[->]             (pr6) -- (pr7);
  \draw[->]             (pr7) -- (pr8);
  \draw[->] (pr8.east) -- ++(2.6,0) -- ++(0,11.2) -- ++(0,2) --                (pr2.east);
  \draw[->]             (pr8) -- (stop);

     \end{tikzpicture}
     \end{center}
\caption{Tri-CgPGA Algorithm Workflow}
\label{fig:foa}
\end{figure}

\begin{algorithm}[H]
\DontPrintSemicolon
\KwIn{Expression Matrix}
\KwOut{Coherent Triclusters}

  Load the expression matrix\;
  Specify the number of cores to be used in parallel\;
            
    \For{tricluster number I =1 to maximum\_triclusters}{
    	Initialise the initial population \;
    	Evaluate the population\;
    	\For{generation number J=1 to maximum\_generations}{
    		selection of parents\;
    		crossover each parent to generate offsprings\;
    		mutation of generated offsprings \;
    		evaluate the new individuals\;
			select the individuals with better fitness \;
        		\If{migration\_interval =count}{
        		
            choose the best individual of the best deem and replace with the worst individual of the other parallel deems\;
       		 }
       
        }
 
        select best individuals from all deems \;
        select the final individual from best\_indiduals\;
        add final tricluster to output\_set \;
        }
        return output\_set\;
        \caption{Tri-CgPGA Pseudo Code}\label{pseudocgpga}
\end{algorithm}

\section{Experimental results and discussions}
\par
All the computational simulations are performed in general conditions on a multiprocessor machine with 4 processors Intel Core i7 3.60 GHz with 4 GB RAM  and Windows 8.1 64 bit operating system memory. The yeast cell cycle dataset (Saccharomyces cerevisiae)~\cite{spellman1998comprehensive} is used for establishing the efficacy of the proposed algorithm. This dataset contains 6179 genes, 4 conditions, and 14 time points. The experiment is performed on the above mentioned dataset along with its two synthetic versions but only reported for the former.

\subsection{\textbf{List of the Parameters}}\label{ds}
\par
During execution, some parameters have been set up like the crossover probability $P_c$, mutation probability $P_m$, weights: $w_g$ for genes, $w_c$ for conditions and $w_t$ for times, distinction weights: $w_{dg}$, $w_{dc}$ and $w_{dt}$ for genes, conditions and times respectively. The details of them are available in table 1. As the algorithms are designed for gene filtration (to obtain the solution with a minimum number of genes), the value of $w_g$ is set to 0.8 so that maximum number of genes can participate in the solution. While setting up the parameters for the distinction term a higher value is being provided for the genes to cover up as much space as possible in this dimension.

\begin{table}[]
\centering
\caption{Values of the parameters taken during algorithm execution}
\label{par_val}
\begin{tabular}{@{}cccccccc@{}}
\hline
$P_c$  & $P_m$   & $w_g$  & $w_c$  & $w_t$  & $w_{dg}$  &$w_{dc}$  & $w_{dt}$    \\
\hline
                            0.8    & 0.5  &0.8    & 0.1 & 0.1 &1.0  & 0.0 & 0.0
\end{tabular}
% \begin{tabular}{cc}
% \hline
% \textbf{Parameter}     & \textbf{Value}          \\ \hline
% $P_c$                     & 0.8                     \\
% $P_m$                     & 0.5                     \\
% $w_g$                     & 0.8                     \\
% $w_c$                    & 0.1                     \\
% $w_t$                   & 0.1                     \\
% $w_{dg}$                    & 1.0                     \\
% $w_{dc}$                    & 0                       \\
% $w_{dt}$                    & 0                       \\ \hline
% \end{tabular}
\end{table}

\subsection{\textbf{Results on Yeast Dataset}}\label{yds}

The simulation results are analyzed from the perspective of the different generations. Analyzing across different generations, it indicates as the number of generations is increased, the values also increase. So for bigger generations, better homogeneity among the genes is obtained which is presented in the following graphs.
\begin{figure}
    \centering
    \begin{subfigure}[b]{0.45\textwidth}
        \includegraphics[width=\textwidth]{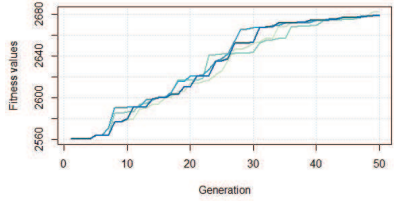}
        \caption{50 Generations}
        \label{fig:50gen}
    \end{subfigure}
    ~ %add desired spacing between images, e. g. ~, \quad, \qquad, \hfill etc. 
      %(or a blank line to force the subfigure onto a new line)
      \hfill
    \begin{subfigure}[b]{0.45\textwidth}
        \includegraphics[width=\textwidth]{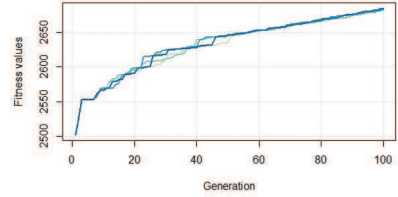}
        \caption{100 Generations}
        \label{fig:100gen}
    \end{subfigure}
    \bigskip
    ~ %add desired spacing between images, e. g. ~, \quad, \qquad, \hfill etc. 
    %(or a blank line to force the subfigure onto a new line)
    \begin{subfigure}[b]{0.45\textwidth}
        \includegraphics[width=\textwidth]{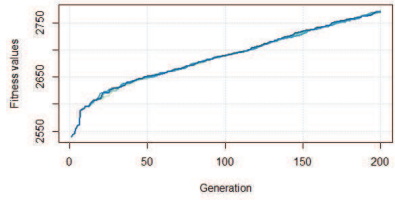}
        \caption{200 Generations}
        \label{fig:200gen}
    \end{subfigure}
    \hfill
    \begin{subfigure}[b]{0.45\textwidth}
        \includegraphics[width=\textwidth]{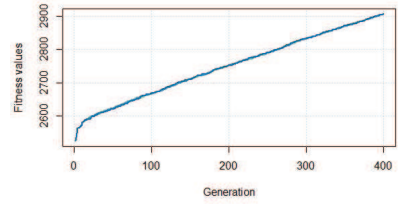}
        \caption{400 Generations}
        \label{fig:400gen}
    \end{subfigure}
    \caption{Fitness Value Plots}\label{fig:fitness}
\end{figure}

\subsection{\textbf{Comparitive Study}}\label{CS}
\par
The results obtained from the execution of the algorithm are quite impressive in terms of time and the volume of the output triclusters. As the fitness function is minimizing, lower the value of  MSR the better is the fitness of the tricluster. Further the results of the Tri-CgPGA algorithm is compared with the results obtained by the trigen algorithm~\cite{gutierrez2014trigen}. The comparison has been done on the basis of computational time taken by the proposed algorithm to execute the codes and to derive the output. In the case of Tri-CgPGA algorithm, it took 30 seconds approximately to run for 1000 genes for 50 generations to deliver the output whereas the trigen algorithm~\cite{gutierrez2014trigen} requires 118 seconds to do the same. Hence exploring parallelism with the genetic approach on triclustering of gene expression microarray data is preferable against the traditional GAs as it reduces the computation time for the algorithm execution. Other relevant information regarding the results obtained from the algorithms Tri-CgPGA algorithm is presented in Table 2.

\setlength{\arrayrulewidth}{0.2mm}
\setlength{\tabcolsep}{2pt}
\renewcommand{\arraystretch}{1}
\begin{table}
\label{table:comparision}
\centering
\caption{Detailed Information about triclusters found by Tri-CgPGA algorithm}
\label{infotable}
\begin{adjustbox}{width=1\textwidth}

\begin{tabular}{@{}cccccc@{}}

 \hline
\textbf{GENE SIZE}     & \textbf{AVG. MSR} & \textbf{AVG. VOLUME} &  \textbf{AVG. NO. OF GENES}   & \textbf{AVG. NO. OF. CONDITIONS} & \textbf{AVG. NO. OF TIME} \\ \hline
\textbf{1000}           & 493.35              & 5124.65                & 616                                  & 1.35                                          & 6.5           \\
\textbf{3000}           & 1322.88             & 33889.5                & 1651.37                              & 3               & 7           \\
\textbf{6178}           & 2669.086            & 67798.75               & 3334                                 & 2.65                                                 & 7.65           \\ \hline
\end{tabular}

\end{adjustbox}

\end{table}
\subsection{\textbf{GO Term Analysis}}\label{go}
\par
The validation of the results obtained is carried out with the Gene Ontology project (GO)~\cite{gene2004gene}. This analysis renders the ontology of terms which describes gene product annotation data along with its characteristics. The ontology describes attributes like molecular functions, cellular component and the relevant biological processes. The queries associated with the associated genes are addressed in GO using the GO Term Finder (Version 0.83)~\cite{boyle2004go}. The findings of the GO Term analysis are presented in Table 3.

\setlength{\arrayrulewidth}{0.2mm}
\setlength{\tabcolsep}{2pt}
\renewcommand{\arraystretch}{1}
\begin{table}[!ht]
\centering
\caption{GO for Yeast Cell Cycle Results}
\label{GOresults}
\begin{adjustbox}{width=1\textwidth}
\begin{tabular}{@{}cllc@{}}
\hline
\textbf{Cluster ID} & \textbf{Biological Process}                                                                              & \textbf{Molecular Function}                                                                                                                                                                                                                                  & \multicolumn{1}{c}{\textbf{\begin{tabular}[c]{@{}c@{}}PI\\ (P-value= \textless{}0.01)\end{tabular}}} \\ \hline
0044699             & Single-Organism Process                                                                 & \begin{tabular}[c]{@{}l@{}}Only one organism is being involved\end{tabular}                                                                                              & 3.02E-10                                                                                             \\ \hline
0016043             & Cellular Component Organization                                                         & \begin{tabular}[c]{@{}l@{}}Assembling or de-assembling of a cellular component constituent parts\end{tabular}                                                           & 4.87E-08                                                                                             \\\hline
0065007             & Biological Regulation                                                                   & \begin{tabular}[c]{@{}l@{}}Biological process regulation of quality or function\end{tabular}                                                                             & 0.00725                                                                                              \\\hline
0080090             & Single-Organism cellular Process                                                        & \begin{tabular}[c]{@{}l@{}}Cellular level activity, occurring within a single organism\end{tabular}                                                                      & 1.61E-06                                                                                             \\\hline
0060255             & Single-Organism Process                                                                 & \begin{tabular}[c]{@{}l@{}}Only one organism is being involved\end{tabular}                                                                                              & 1.91E-06                                                                                             \\\hline
0019222             & Single-Organism Process                                                                 & \begin{tabular}[c]{@{}l@{}}Only one organism is being involved\end{tabular}                                                                                              & 0.00019                                                                                              \\\hline
0044763             & Single-Organism Cellular Process                                                        & \begin{tabular}[c]{@{}l@{}}Cellular level activity, occurring within a single organism\end{tabular}                                                                      & 2.69E-06                                                                                             \\\hline
0050789             & Single-Organism Process                                                                 & \begin{tabular}[c]{@{}l@{}}Only one organism is being involved\end{tabular}                                                                                              & 4.10E-05                                                                                             \\\hline
2000112             & Single-Organism Cellular Process                                                        & \begin{tabular}[c]{@{}l@{}}Cellular level activity, occurring within a single organism\end{tabular}                                                                      & 4.89E-06                                                                                             \\\hline
0010556             & Single-Organism Cellular Process                                                        & \begin{tabular}[c]{@{}l@{}}Cellular level activity, occurring within a single organism\end{tabular}                                                                      & 0.00315                                                                                              \\\hline
0071840             & Cellular Component Organization                                                         & \begin{tabular}[c]{@{}l@{}}Biosynthesis of constituent macromolecules,assembly, arrangement of\\ constituent parts, or disassembly of a cellular component\end{tabular} & 0.00753                                                                                              \\\hline
0051171             & \begin{tabular}[c]{@{}l@{}}Cellular Component Organization or\\ Biogenesis\end{tabular} & \begin{tabular}[c]{@{}l@{}}Biosynthesis of constituent macromolecules, assembly, arrangement of\\ constituent parts, or disassembly of a cellular component\end{tabular} & 0.00026                                                                                              \\\hline
0006996             & Organelle Organization                                                                  & \begin{tabular}[c]{@{}l@{}}Cellular level assembly, arrangement of constituent parts,\\ or disassembly of an organelle within a cell\end{tabular}                        & 6.00541                                                                                              \\\hline
0010468             & Organelle Organization                                                                  & \begin{tabular}[c]{@{}l@{}}Cellular level assembly, arrangement of constituent parts,\\ or disassembly of an organelle within a cell\end{tabular}                        & 0.00563                                                                                              \\\hline
0032774             & Biological Regulation                                                                   & \begin{tabular}[c]{@{}l@{}}Biological process regulation of quality or function\end{tabular}                                                                             & 0.00939                                                                                             
 \\ \hline 
\end{tabular}
\end{adjustbox}
\end{table}

\section{Conclusion}\label{conc}
\par 
A new framework Tri-CgPGA, based on the coarse-grained parallel genetic approach(CgPGA) to generate the triclusters from gene expression database is proposed in our work. The results of the suggested framework are compared with another state of the art technique called as Trigen algorithm. As the comparison justifies the proposed scheme's efficiency over the existing schemes considering the computation time, hence it is preferable to adopt parallel GAs over traditional GAs in the triclustering of gene expression 3D microarray data. There exist number of  future directions which might further improve this framework: (1) The acquisition of large-scale databases from other standard datasets to measure the performance of the frameworks (2) To further improve the coherence and the computation time, other competent evaluation measures with the suggested or other existing versions of PGAs should be investigated to obtain more meaningful triclusters.\\

\bibliographystyle{splncs04}
\renewcommand{\bibname}{References}
\bibliography{main}

\end{document}